\documentclass{article}
\usepackage[table]{xcolor}
\usepackage{arxiv,times}
\iclrfinalcopy
\usepackage{tablefootnote}
\usepackage{refcount}

\usepackage{subcaption}
\usepackage{amsmath,amssymb,graphicx,color,algorithm, algpseudocode,booktabs,bm,relsize,enumitem,multirow,amsthm,epsfig,caption,bm,capt-of}
\usepackage{url}
\usepackage{wrapfig}

\renewcommand{\headrulewidth}{0pt}

\newsavebox{\bigleftbox}

\renewcommand{\thefootnote}{\alph{footnote}}

\newcommand{\astfootnote}[1]{%
\let\oldthefootnote=\thefootnote%
\setcounter{footnote}{0}%
\renewcommand{\thefootnote}{\fnsymbol{footnote}}%
\footnote{#1}%
\let\thefootnote=\oldthefootnote%
}

\definecolor{Blue9}{rgb}{0.1,0.3,0.95}
\usepackage[colorlinks=true,linkcolor=Blue9,citecolor=Blue9]{hyperref}

\title{\vspace{-0.15in}HARP: Autoregressive Latent Video Prediction with High-Fidelity Image Generator\vspace{-0.1in}}

\author{
  Younggyo Seo$^{1,*}$\hspace{-0.03in}
  \,
  Kimin Lee$^{2,\dagger}$\hspace{-0.03in}
  \,
  Fangchen Liu$^{2}$\hspace{-0.03in}
  \,
  Stephen James$^{2,\ddagger}$\hspace{-0.03in}
  \,
  Pieter Abbeel$^{2}$\hspace{-0.03in}
  \\
  $^{1}$KAIST\;
  $^{2}$UC Berkeley
  \vspace{-0.2in}
}

\begin{document}

\maketitle

\begin{abstract}
Video prediction is an important yet challenging problem; burdened with the tasks of generating future frames and learning environment dynamics.
Recently, autoregressive latent video models have proved to be a powerful video prediction tool, by separating the video prediction into two sub-problems: pre-training an image generator model, followed by learning an autoregressive prediction model in the latent space of the image generator.
However, successfully generating high-fidelity and high-resolution videos has yet to be seen.
In this work, we investigate how to train an autoregressive latent video prediction model capable of predicting high-fidelity future frames with minimal modification to existing models, and produce high-resolution (256x256) videos.
Specifically, we scale up prior models by employing a high-fidelity image generator (VQ-GAN) with a causal transformer model, and introduce additional techniques of top-$k$ sampling and data augmentation to further improve video prediction quality.
Despite the simplicity, the proposed method achieves competitive performance to state-of-the-art approaches on standard video prediction benchmarks with fewer parameters, and enables high-resolution video prediction on complex and large-scale datasets.
Videos are available at \url{https://sites.google.com/view/harp-videos/home}\astfootnote{Work done while visiting UC Berkeley \; $^\dagger$Now at Google Research \; $^\ddagger$Now at Dyson Robot Learning Lab \vspace{0.03in}\\ © 2022 IEEE.  Personal use of this material is permitted.  Permission from IEEE must be obtained for all other uses, in any current or future media, including reprinting/republishing this material for advertising or promotional purposes, creating new collective works, for resale or redistribution to servers or lists, or reuse of any copyrighted component of this work in other works.}.
\end{abstract}

\vspace{-0.1in}
\section{Introduction}
\label{sec:intro}
\begin{figure}[h]
\vspace{-0.225in}
\centering
\includegraphics[width=0.975\linewidth]{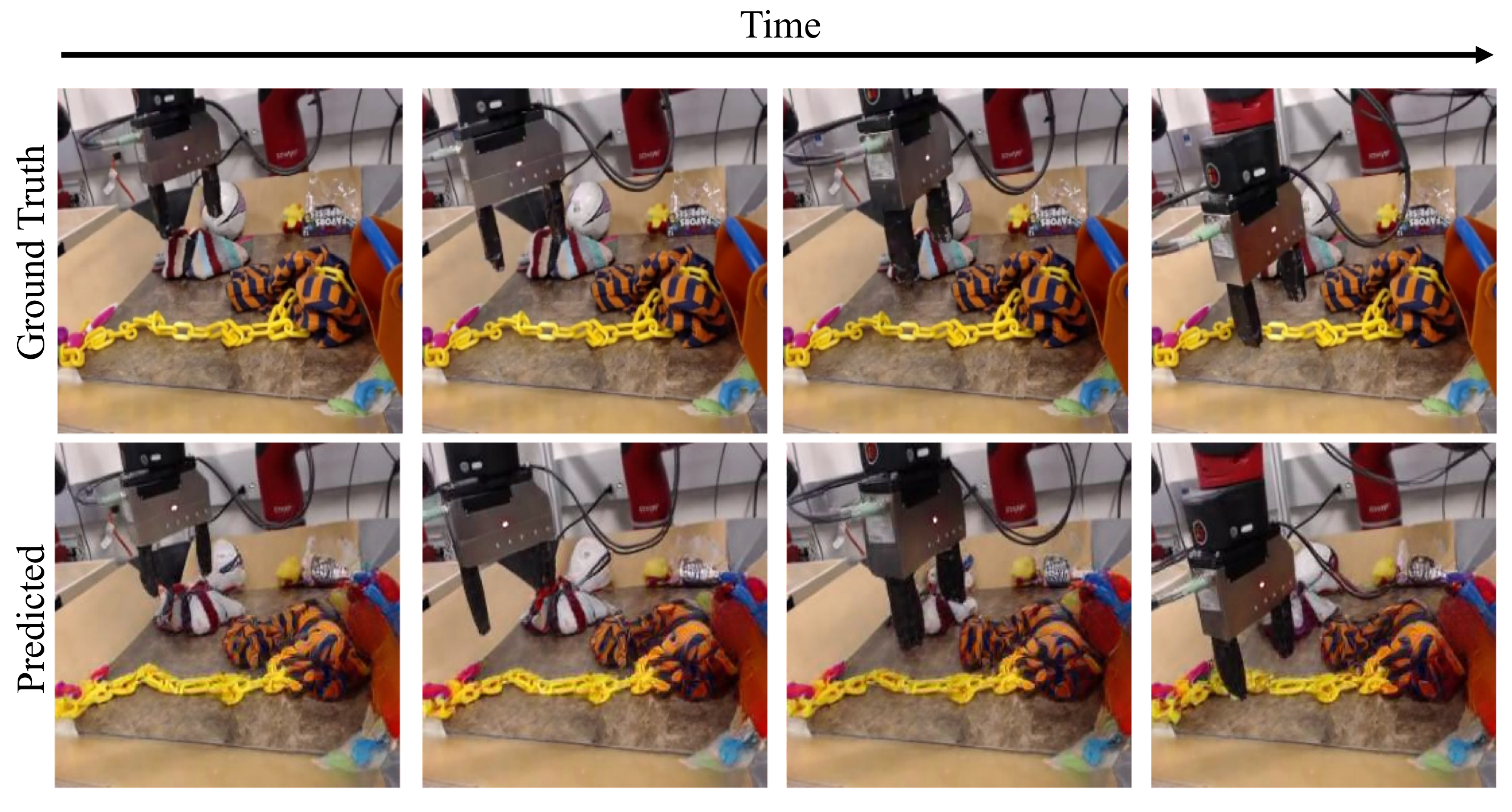}
\vspace{-0.15in}
\caption{Selcted $256 \times 256$ video sample generated by HARP on RoboNet \citep{dasari2019robonet}.}
\label{fig:robonet_intro}
\end{figure}
Video prediction can enable agents to learn useful representations for predicting the future consequences of the decisions they make,
which is crucial for solving the tasks that require long-term planning, including robotic manipulation \citep{finn2017deep,kalashnikov2018qt} and autonomous driving \citep{xu2017end}.
Despite the recent advances in improving the quality of video prediction \citep{finn2016unsupervised,babaeizadeh2017stochastic,denton2018stochastic,lee2018stochastic,weissenborn2019scaling,babaeizadeh2021fitvid}, learning an accurate video prediction model remains notoriously difficult problem and requires a lot of computing resources, especially when the inputs are video sequences with high-resolution \citep{villegas2019high,clark2019adversarial,luc2020transformation}.
This is because the video prediction model should excel at both tasks of generating high-fidelity images and learning the dynamics of environments, though each task itself is already a very challenging problem.

Recently, autoregressive latent video prediction methods \citep{rakhimov2020latent,yan2021videogpt,yan2022patch} have been proposed to improve the efficiency of video prediction, by separating video prediction into two sub-problems: first pre-training an image generator (\textit{e.g.,} VQ-VAE; \citealt{oord2017neural}), and then learning the autoregressive prediction model \citep{weissenborn2019scaling, chen2020generative} in the latent space of the pre-trained image generator.
However, the prior works are limited in that they only consider relatively low-resolution videos (up to $128 \times 128$ pixels) for demonstrating the efficiency of the approach; it is questionable that such experiments can fully demonstrate the benefit of operating in the latent space of image generator instead of pixel-channel space.

In this paper, we present \textbf{H}igh-fidelity \textbf{A}uto\textbf{R}egressive latent video \textbf{P}rediction (HARP), which scales up the previous autoregressive latent video prediction methods for high-fidelity video prediction. The main principle for the design of HARP is simplicity: we improve the video prediction quality with minimal modification to existing methods.
First, for image generation, we employ a high-fidelity image generator, \textit{i.e.,} vector-quantized generative adversarial network (VQ-GAN; \citealt{esser2021taming}). This improves video prediction by enabling high-fidelity image generation (up to $256 \times 256$ pixels) on various video datasets.
Then a causal transformer model \citep{chen2020generative}, which operates on top of discrete latent codes, is trained to predict the discrete codes from VQ-GAN, and autoregressive predictions made by the transformer model are decoded into future frames at inference time.

We highlight the main contributions of this paper below:
\begin{itemize}
    \item [$\bullet$] We show that our autoregressive latent video prediction model, HARP, can predict high-resolution ($256 \times 256$ pixels) frames on robotics dataset (\textit{i.e.,} Meta-World \citep{yu2020meta}) and large-scale real-world robotics dataset (\textit{i.e.,} RoboNet \citep{dasari2019robonet}).
    \item [$\bullet$] We show that HARP can leverage the image generator pre-trained on ImageNet for training a high-resolution video prediction model on complex, large-scale Kinetics-600 dataset \citep{carreira2018short}.
    \item [$\bullet$] HARP achieves competitive or superior performance to prior state-of-the-art video prediction models on widely-used BAIR Robot Pushing \citep{ebert2017self} and KITTI driving \citep{geiger2013vision} video prediction benchmarks.
\end{itemize}

\section{Related work}
\paragraph{Video prediction.} Video prediction aims to predict the future frames conditioned on images \citep{michalski2014modeling, ranzato2014video,srivastava2015unsupervised,vondrick2016generating,lotter2016deep}, texts \citep{wu2021godiva}, and actions \citep{oh2015action, finn2016unsupervised}, which would be useful for several applications, \textit{e.g.,} model-based RL \citep{hafner2019learning,kaiser2019model,hafner2020mastering,rybkin2021model,seo2022masked,seo2022reinforcement}, and simulator development \citep{kim2020learning,kim2021drivegan}.
Various video prediction models have been proposed with different approaches, including generative adversarial networks (GANs; \citealt{goodfellow2014generative}) known to generate high-fidelity images by introducing adversarial discriminators that also considers temporal or motion information \citep{aigner2018futuregan,jang2018video,kwon2019predicting,clark2019adversarial,luc2020transformation,skorokhodov2022stylegan,yu2022generating}, latent video prediction models that operates on the latent space \citep{babaeizadeh2017stochastic, denton2018stochastic, lee2018stochastic, villegas2019high, wu2021greedy, babaeizadeh2021fitvid}, and autoregressive video prediction models that operates on pixel space by predicting the next pixels in an autoregressive way \citep{kalchbrenner2017video,reed2017parallel,weissenborn2019scaling}.
\clearpage

\paragraph{Autoregressive latent video prediction.} Most closely related to our work are autoregressive latent video prediction models that separate the video prediction problem into image generation and dynamics learning.
\citet{walker2021predicting} proposed to learn a hierarchical VQ-VAE \citep{razavi2019generating} that extracts multi-scale hierarchical latents then train SNAIL blocks \citep{chen2018pixelsnail} that predict hierarchical latent codes, enabling high-fidelity video prediction. However, this involves a complicated training pipeline and a video-specific architecture, which limits its applicability.
As simple alternatives, \citet{rakhimov2020latent,yan2021videogpt,yan2022patch} proposed to first learn a VQ-VAE \citep{oord2017neural} and train a causal transformer with 3D self-attention \citep{weissenborn2019scaling} and factorized 2D self-attention \citep{child2019generating}, respectively. These approaches, however, are limited in that they only consider low-resolution videos.
We instead present a simple high-resolution video prediction method that incorporates the strengths of both prior approaches.

\begin{figure}[t!]
  \centering
  \includegraphics[width=0.99\linewidth]{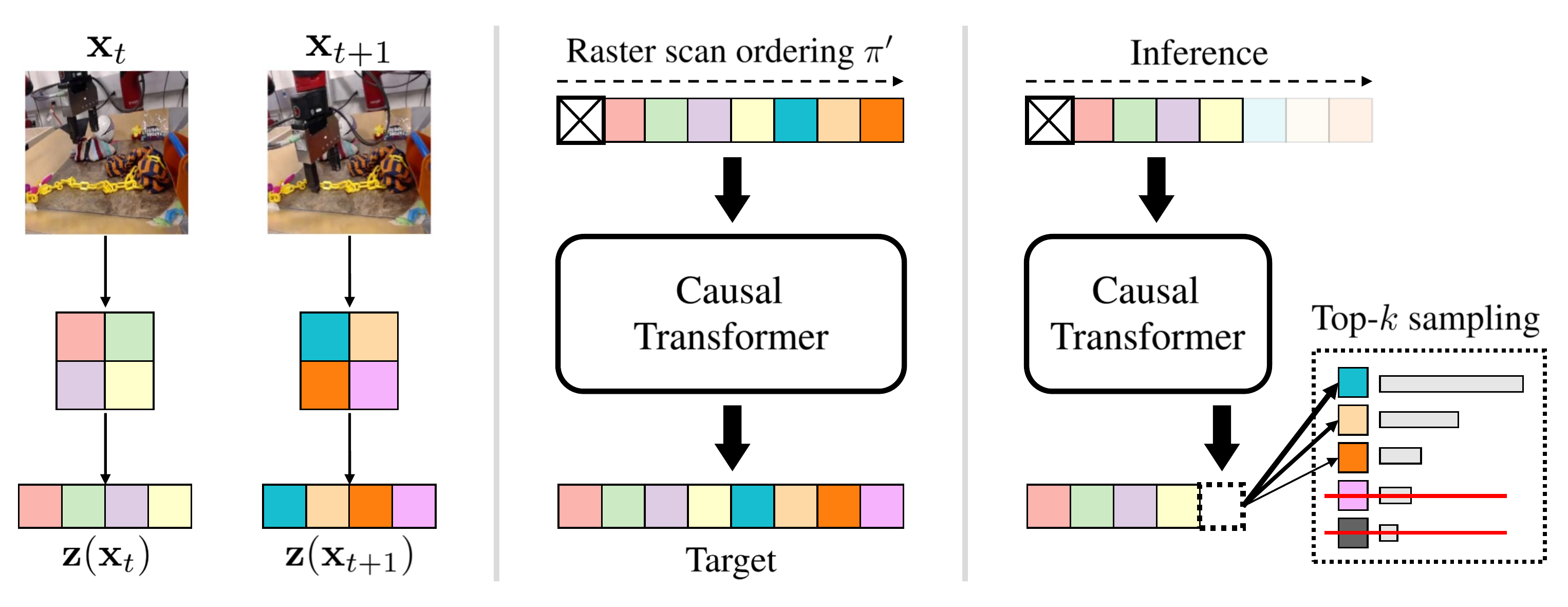}
  \vspace{-0.15in}
\caption{Illustration of our approach. We first train a VQ-GAN model that encodes frames into discrete latent codes. Then the discrete codes are flattened following the raster scan order, and a causal transformer model is trained to predict the next discrete codes in an autoregressive manner.}
\label{fig:overview}
\vspace{-0.05in}
\end{figure}

\section{Preliminaries}
\label{sec:preliminaries}
We aim to learn a video prediction model that predicts the future frames $\mathbf{x}_{c:T} = (\mathbf{x}_{c}, ..., \mathbf{x}_{T-1})$ conditioned on the first $c$ frames of a video $\mathbf{x}_{<c} = (\mathbf{x}_{0}, \mathbf{x}_{1}, ..., \mathbf{x}_{c-1})$, where $\mathbf{x}_{t} \in \mathbb{R}^{H \times W \times N_{ch}}$ is the frame at timestep $t$.
Optionally, one can also consider conditioning the prediction model on actions $\mathbf{a} = (\mathbf{a}_{0}, ..., \mathbf{a}_{T-1})$ that the agents would take.

\subsection{Autoregressive video prediction model}
Autoregressive video prediction model \citep{weissenborn2019scaling} approximates the distribution of a video in a pixel-channel space. Given a video $\mathbf{x} \in \mathbb{R}^{T \times H \times W \times N_{ch}}$, the joint distribution over pixels conditioned on the first $c$ frames is modelled as the product of channel intensities $N_{ch}$ and all $N_{p}=T \cdot H \cdot W$ pixels except $N_{c} = c \cdot H \cdot W$ pixels of conditioning frames:
\begin{align}\label{eq:autoregressive}
    p(\mathbf{x}_{c:T}\,|\,\mathbf{x}_{<c}) = \prod^{N_{p}-1}_{i=N_{c} - 1} \prod^{N_{ch}-1}_{k=0} p(\mathbf{x}_{\pi(i)}^{k} | \mathbf{x}_{\pi(<i)}, \mathbf{x}_{\pi(i)}^{<k}),
\end{align}
where $\pi$ is a raster-scan ordering over all pixels from the video (we refer to \citet{weissenborn2019scaling} for more details), $\mathbf{x}_{\pi(<i)}$ is all pixels before $\mathbf{x}_{\pi(i)}$, $\mathbf{x}_{\pi(i)}^{k}$ is $k$-th channel intensity of the pixel $\mathbf{x}_{\pi(i)}$, and $\mathbf{x}_{\pi(i)}^{<k}$ is all channel intensities before $\mathbf{x}_{\pi(i)}^{k}$.

\subsection{Vector quantized variational autoencoder}
VQ-VAE \citep{oord2017neural} consists of an encoder that compresses images into discrete representations, and a decoder that reconstructs images from these discrete representations.
Formally, given an image $x \in \mathbb{R}^{H \times W \times N_{ch}}$, the encoder $E$ encodes $x$ into a feature map $z_{e}(x) \in \mathbb{R}^{H' \times W' \times N_{z}}$ consisting of a series of latent vectors $z_{\pi'(i)}(x) \in \mathbb{R}^{N_{z}}$, where $\pi'$ is a raster-scan ordering of the feature map $z_{e}(x)$ of size $|\pi'| = H' \cdot W'$. Then $z_{e}(x)$ is quantized to discrete representations $z_{q}(x) \in \mathbb{R}^{|\pi'| \times N_{z}}$ based on the distance of latent vectors $z_{\pi'(i)}(x)$ to the prototype vectors in a codebook $C = \{e_{k}\}_{k=1}^{K}$ as follows:
\begin{align}\label{eq:vq}
    &z_{q}(x) = (e_{q(x, 1)}, e_{q(x, 2)}, \cdots, e_{q(x, |\pi'|)}),\nonumber\\ &\text{where}\; q(x, i)=\text{argmin}_{k \in [K]} \Vert z_{\pi'(i)}(x) - e_{k}\Vert_{2},
\end{align}
where $[K]$ is the set $\{1, \cdots, K\}$. Then the decoder $G$ learns to reconstruct $x$ from discrete representations $z_{q}(x)$. The VQ-VAE is trained by minimizing the following objective:
\begin{align}\label{eq:vqvaeloss}
    \mathcal{L}_{\tt{VQVAE}}(x) &= \underbrace{\Vert x - G(z_{q}(x))\Vert_{2}^{2}}_{\mathcal{L}_{\tt{recon}}} + \underbrace{\Vert sg\left[z_{e}(x)\right] - z_{q}(x) \Vert_{2}^{2}}_{\mathcal{L}_{\tt{codebook}}} + \underbrace{\beta \cdot \Vert sg\left[z_{q}(x)\right] - z_{e}(x) \Vert_{2}^{2}}_{\mathcal{L}_{\tt{commit}}},
\end{align}
where the operator $sg$ refers to a stop-gradient operator, $\mathcal{L}_{\tt{recon}}$ is a reconstruction loss for learning representations useful for reconstructing images, $\mathcal{L}_{\tt{codebook}}$ is a codebook loss to bring codebook representations closer to corresponding encoder outputs $h$, and $\mathcal{L}_{\tt{commit}}$ is a commitment loss weighted by $\beta$ to prevent encoder outputs from fluctuating frequently between different representations.

\subsection{Vector quantized generative adversarial network} VQ-GAN \citep{esser2021taming} is a variant of VQ-VAE that (a) replaces the $\mathcal{L}_{\tt recon}$ in (\ref{eq:vqvaeloss}) by a perceptual loss $\mathcal{L}_{\tt{LPIPS}}$ \citep{zhang2018unreasonable}, and (b) introduces an adversarial training scheme where a patch-level discriminator $D$ \citep{isola2017image} is trained to discriminate real and generated images by maximizing following loss:
\begin{align}
    \mathcal{L}_{\tt{GAN}}(x) = [\log D(x) + \log (1 - D(G(z_{q}(x)))].
\end{align}
Then, the objective is given as below:
\begin{align}
    \min_{E, G, C} \max_{D} \mathbb{E}_{x \sim p(x)} \big[ \big( &\mathcal{L_{\tt{LPIPS}}} + \mathcal{L_{\tt{codebook}}} + \mathcal{L_{\tt{commit}}}\big) + \lambda \cdot \mathcal{L_{\tt{GAN}}} \big],
\end{align}
where $\lambda = \frac{\nabla_{G_{L}}[\mathcal{L}_{\tt{LPIPS}}]}{\nabla_{G_{L}}[\mathcal{L}_{\tt{GAN}}] + \delta}$ is an adaptive weight, $\nabla_{G_{L}}$ is the gradient of the inputs to the last layer of the decoder $G_{L}$, and $\delta = 10^{-6}$ is a scalar introduced for numerical stability.

\section{Method}
We present HARP, a video prediction model capable of predicting high-fidelity future frames. Our method is designed to fully exploit the benefit of autoregressive latent video prediction model that separates the video prediction into image generation and dynamics learning.
The full architecture of HARP is illustrated in Figure~\ref{fig:overview}.

\subsection{High-fidelity image generator}
We utilize the VQ-GAN model \citep{esser2021taming} that has proven to be effective for high-resolution image generation as our image generator (see Section~\ref{sec:preliminaries} for the formulation of VQ-GAN).
Specifically, we first pre-train the image generator then freeze the model throughout training to improve the efficiency of learning video prediction models.
The notable difference to a prior work that utilize 3D convolutions to temporally downsample the video for efficiency \citep{yan2021videogpt} is that our image generator operates on single images; hence our image generator solely focus on improving the quality of generated images.
Importantly, 
this enables us to utilize the VQ-GAN model pre-trained on a wide range of natural images, \textit{e.g.,} ImageNet, without training the image generator on the target datasets, which can significantly reduce the training cost of high-resolution video prediction model.

\subsection{Autoregressive latent video prediction model}
To leverage the VQ-GAN model for video prediction, we utilize the autoregressive latent video prediction architecture that operates on top of the discrete codes. Specifically, we extract the discrete codes $\mathbf{z(\mathbf{x})} = (\mathbf{z}(\mathbf{x}_{1}), ..., \mathbf{z}(\mathbf{x}_{T}))$ using the pre-trained VQ-GAN, where $\mathbf{z}(\mathbf{x}_{t}) = (q_{(\mathbf{x}_{t}, 1)}, q_{(\mathbf{x}_{t}, 2)}, ..., q_{(\mathbf{x}_{t}, |\pi'|)})$ is the discrete code extracted from the frame $\mathbf{x}_{t}$ as in (\ref{eq:vq}). Then, instead of modelling the distribution of video $p(\mathbf{x})$ in the pixel-channel space as in (\ref{eq:autoregressive}), we learn the distribution of the video in the discrete latent representation space:
\begin{align}\label{eq:latent_autoregressive}
    p(\mathbf{z}(\mathbf{x}_{c:T} | \mathbf{x}_{<c})) = \prod^{N_{d}-1}_{i=0} p(\mathbf{z}_{\pi'(i)}(\mathbf{x}) | \mathbf{z}_{\pi'(<i)}(\mathbf{x})),
\end{align}
where $N_{d} = (T - C) \cdot H' \cdot W'$ is the total number of codes from $\mathbf{x}_{c:T}$.
Due to its simplicity, we utilize the causal transformer architecture \citep{yan2021videogpt} where the output logits from input codes are trained to predict the next discrete codes.

\subsection{Additional techniques}

\paragraph{Top-$\textbf{k}$ sampling.}
To improve the video prediction quality of latent autoregressive models whose outputs are sampled from the probability distribution over a large number of discrete codes, we utilize the top-$k$ sampling \citep{fan2018hierarchical} that randomly samples the output from the top-$k$ probable discrete codes.
By preventing the model from sampling rare discrete codes from the long-tail of a probability distribution and predicting future frames conditioned on such discrete codes, we find that top-$k$ sampling improves video prediction quality, especially given that the number of discrete encodings required for future prediction is very large, \textit{e.g.,} 2,560 on RoboNet \citep{dasari2019robonet} up to 6,400 on KITTI dataset \citep{geiger2013vision} in our experimental setup. 

\paragraph{Data augmentation.}
We also investigate how data augmentation can be useful for improving the performance of autoregressive latent video prediction models. Since the image generator model is not trained with augmentation, we utilize a weak augmentation to avoid the instability coming from aggressive transformation of input frames, \textit{i.e.,} translation augmentation that moves the input images by $m$ pixels along the X or Y direction.

\section{Experiments}
\label{sec:experiments}
We design our experiments to investigate the following:
\begin{itemize}
    \item [$\bullet$] Can HARP predict high-resolution future frames (up to $256 \times 256$ pixels) on various video datasets with different characteristics?
    \item [$\bullet$] How does HARP compare to state-of-the-art methods with large end-to-end networks on standard video prediction benchmarks in terms of quantitative evaluation?
    \item [$\bullet$] How does the proposed techniques affect the performance of HARP?
\end{itemize}

\begin{figure*}[t!] \centering
\subfloat[RoboNet]
{
\includegraphics[width=0.485\textwidth]{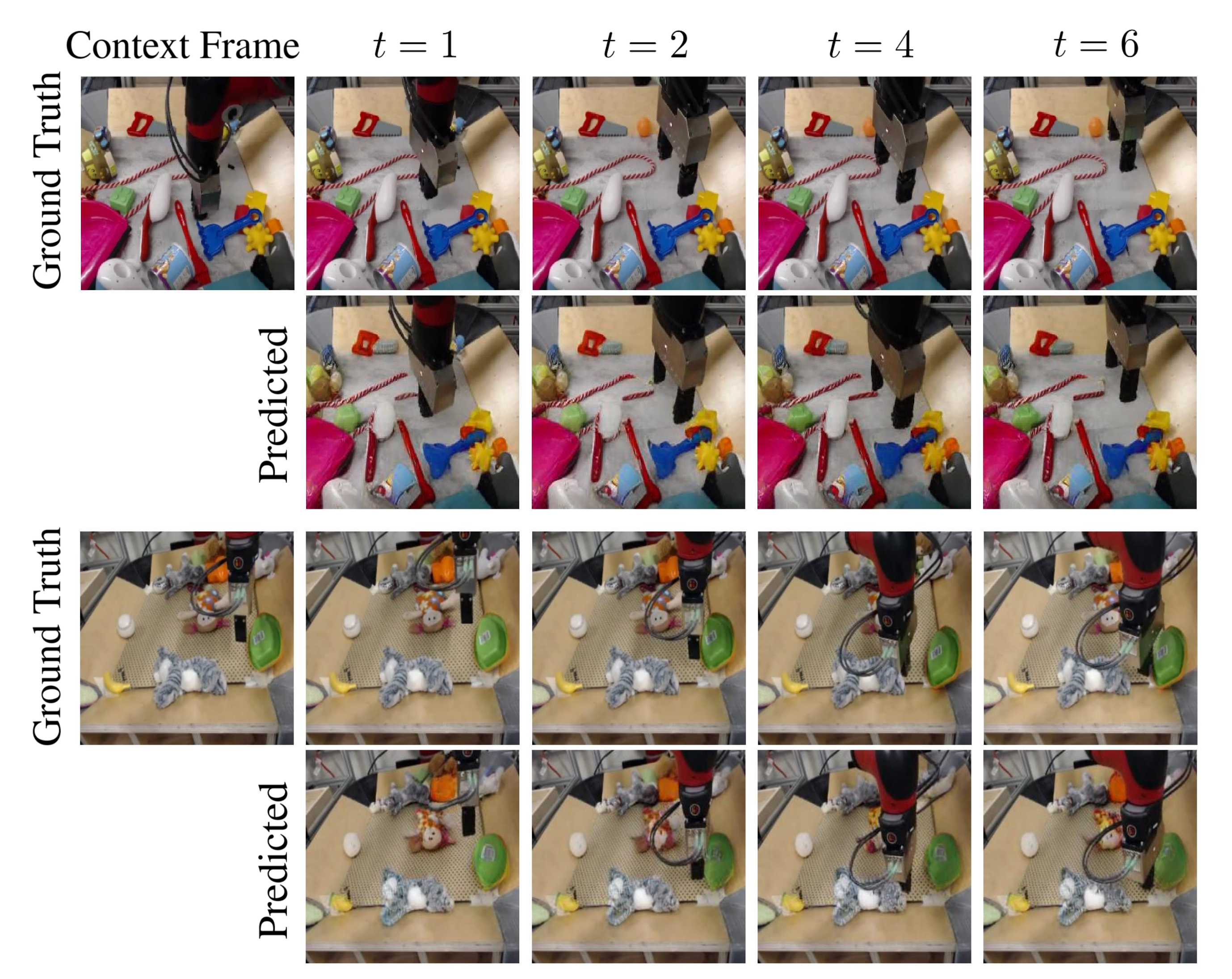}
\label{fig:video_prediction_robonet}}
\subfloat[Kinetics-600]
{
\includegraphics[width=0.485\textwidth]{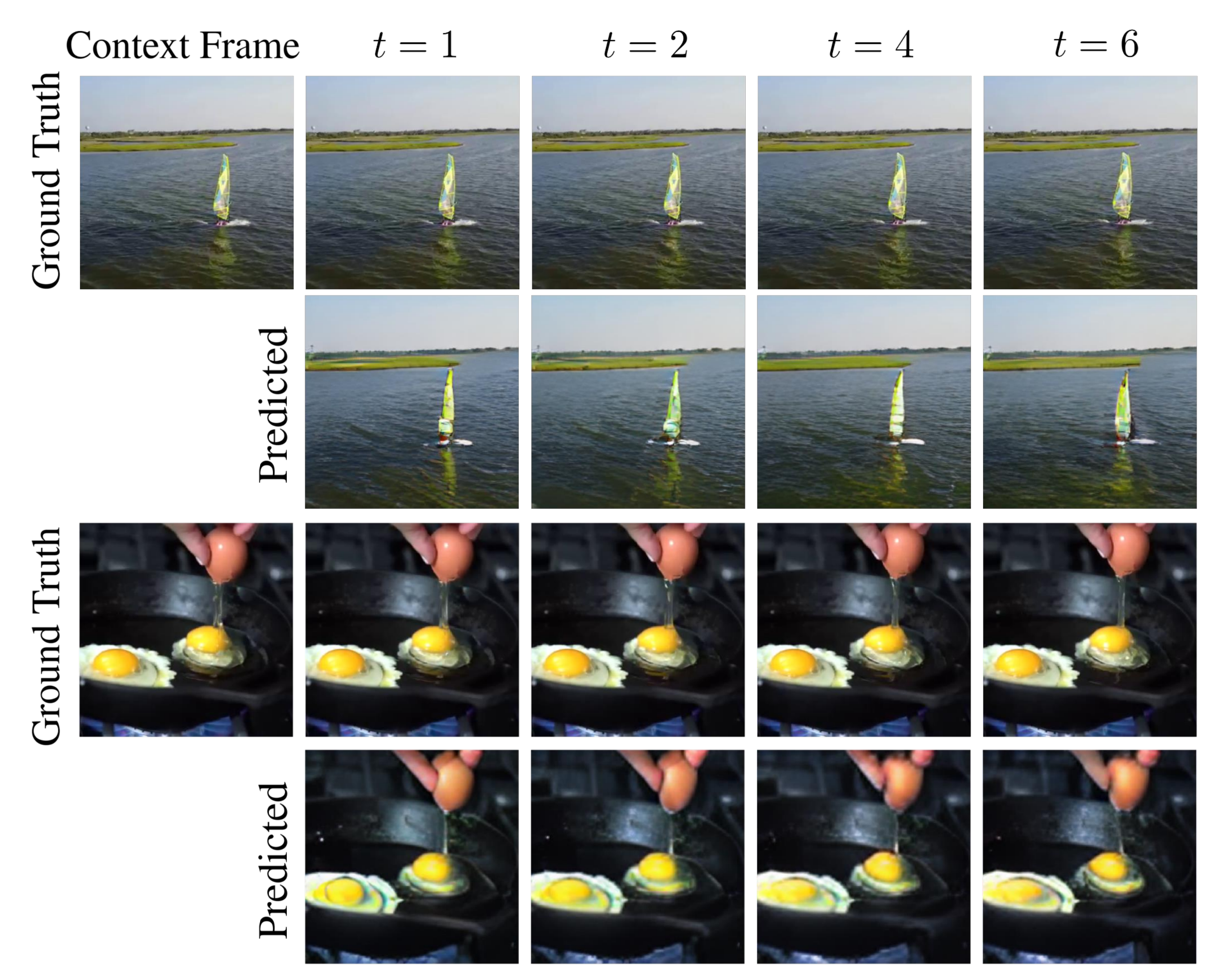}
\label{fig:video_prediction_kinetics}}
\vspace{-0.1in}
\caption{$256 \times 256$ future frames predicted by HARP trained on (a) RoboNet \citep{dasari2019robonet} and (b) Kinetics-600 \citep{carreira2018short} datasets.}
\label{fig:video_prediction}
\end{figure*}

\begin{table}[t!]
  \caption{Quantitative evaluation on (a) BAIR Robot Pushing \citep{ebert2017self} and (b) KITTI driving dataset \citep{geiger2013vision}. We observe that HARP can achieve competitive performance to state-of-the-art methods with large end-to-end networks on these benchmarks.}
  \hfill
  \subfloat[BAIR Robot Pushing]
  {
  \small
  \begin{tabular}{lrr}\toprule
    Method\tablefootnote{
    Baselines are SVG \citep{villegas2019high}, GHVAE \citep{wu2021greedy}, FitVid \citep{babaeizadeh2021fitvid}, LVT \citep{rakhimov2020latent}, SAVP \citep{lee2018stochastic}, DVD-GAN-FP \citep{clark2019adversarial}, VideoGPT \citep{yan2021videogpt}, TrIVD-GAN-FP \citep{luc2020transformation}, and Video Transformer \citep{weissenborn2019scaling}.
    } &Params &FVD ($\downarrow$)\\\midrule
    LVT &50M &125.8 \\
    SAVP &53M &116.4 \\
    DVD-GAN-FP &---$^\dagger$ &109.8 \\
    VideoGPT &82M &103.3 \\
    TrIVD-GAN-FP &---$^\dagger$ &103.3 \\
    Video Transformer &373M &94.0 \\
    FitVid &302M &\textbf{93.6} \\\midrule
    HARP (ours) &89M &99.3 \\
    \bottomrule
    \end{tabular}
  }
  \subfloat[KITTI]
  {
  \small
  \begin{tabular}{lrrr}\toprule
    Method$^{\text{\textcolor{Blue9}{4}}}$ &Params &FVD ($\downarrow$) &LPIPS ($\downarrow$)\\\midrule
    SVG &298M &1217.3 &0.327 \\
    GHVAE &599M  &552.9 &0.286 \\
    FitVid &302M &884.5 &0.217\\\midrule
    HARP (ours) &89M &\textbf{482.9} & \textbf{0.191}\\
    \bottomrule
    \noalign{\vspace{3pt}}
    \raggedright\footnotesize $^{\dagger}$ Not available
    \end{tabular}
  }
  \hfill
  \label{tbl:quantitative}
\end{table}

\subsection{High-resolution video prediction}
\paragraph{Implementation.} We utilize up to 8 Nvidia 2080Ti GPU and 20 CPU cores for training each model.
For training VQ-GAN \citep{esser2021taming}, we first train the model without a discriminator loss $\mathcal{L}_{\tt{GAN}}$, and then continue the training with the loss following the suggestion of the authors.
For all experiments, VQ-GAN downsamples each frame into $16 \times 16$ latent codes, \textit{i.e.,} by a factor of 4 for frames of size $64 \times 64$ frames, and 16 for frames of size $256 \times 256$.
For training a transformer model, the VQ-GAN model is frozen so that its parameters are not updated.
We use Sparse Transformers \citep{child2019generating} as our transformer architecture to accelerate the training.
For hyperparameterse, we use $k = 10$ for sampling at inference time.

\paragraph{Setup.} For all experiments, VQ-GAN downsamples each frame into $16 \times 16$ latent codes, \textit{i.e.,} by a factor of 4 for frames of size $64 \times 64$ frames, and 16 for frames of size $256 \times 256$.
For training a transformer model, the VQ-GAN model is frozen so that its parameters are not updated.
As for hyperparameter, we use $k = 10$ for sampling at inference time, but no data augmentation for high-resolution video prediction experiments.
We investigate how our model works on large-scale real-world RoboNet dataset \citep{dasari2019robonet} consisting of more than 15 million frames, and Kinetics-600 dataset consisting of more than 400,000 videos, which require a large amount of computing resources for training even on $64 \times 64$ resolution \citep{babaeizadeh2021fitvid,clark2019adversarial}.
For RoboNet experiments, we first train a VQ-GAN model, and then train a 12-layer causal transformer model that predicts future 10 frames conditioned on first two frames and future ten actions.
For Kinetics-600 dataset, to avoid the prohibitively expensive training cost of high-resolution video prediction models on this dataset and fully exploit the benefit of employing a high-fidelity image generator, we utilize the ImageNet pre-trained VQ-GAN model.
As we train the transformer model only for autoregressive prediction, this enables us to train a video prediction model in a very efficient manner. 

\paragraph{Results.} First, we provide the predicted frames on the held-out test video of RoboNet dataset in Figure~\ref{fig:video_prediction_robonet}, where the model predicts the high-resolution future frames where a robot arm is moving around various objects of different colors and shapes.
Furthermore, Figure~\ref{fig:video_prediction_kinetics} shows that Kinetics-600 pre-trained model can also predict future frames on the test natural videos\footnote{Videos with CC-BY license: Figure~\ref{fig:video_prediction_kinetics} \href{https://www.youtube.com/watch?v=p9f3BPInhLI
}{top} and \href{https://www.youtube.com/watch?v=p9f3BPInhLI
}{bottom}}, which demonstrates that leveraging the large image generator pre-trained on a wide range of natural images can be a promising recipe for efficient video prediction on large-scale video datasets. 

\subsection{Comparative evaluation on standard benchmarks}
\paragraph{Setup.} For quantitative evaluation, we first consider the BAIR robot pushing dataset \citep{ebert2017self} consisting of roughly 40k training and 256 test videos. Following the setup in prior work \citep{yan2021videogpt}, we predict 15 future frames conditioned on one frame. We also evaluate our method on KITTI driving dataset \citep{geiger2013vision}, where the training and test datasets are split by following the setup in \citet{villegas2019high}.
Specifically, the test dataset consists of 148 video clips constructed by extracting 30-frame clips and skipping every 5 frames, and the model is trained to predict future ten frames conditioned on five frames and evaluated to predict future 25 frames conditioned on five frames.
For hyperparameters, We use $k$ = 10 for both datasets and data augmentation with $m$ = 4 is only applied to KITTI as there was no sign of overfitting on BAIR dataset.
For evaluation metrics, we use LPIPS \citep{zhang2018unreasonable} and FVD \citep{unterthiner2018towards}, computed using 100 future videos for each ground-truth test video, then reports the best score over 100 videos for LPIPS, and all videos for FVD, following \citet{babaeizadeh2021fitvid,villegas2019high}.

\begin{table}[t!]
  \centering\small
  \caption{FVD scores of HARP with varying (a) the number of codes to use for top-$k$ sampling, (b) number of layers, and (c) magnitude $m$ of data augmentation. }
  \vspace{-0.1in}
  \hfill
  \begin{minipage}{.33\linewidth}
  \centering
  \subfloat[Effects of $k$]
  {
    \begin{tabular}{ccr}
    \toprule
     Dataset & \multicolumn{1}{c}{$k$} & \multicolumn{1}{c}{FVD ($\downarrow$)} \\ 
     \midrule
     \multirow{3}{*}{BAIR} & No top-$k$ & \multicolumn{1}{r}{104.4} \\
      & 100 & \multicolumn{1}{r}{103.6} \\
      & 10 & \multicolumn{1}{r}{\textbf{99.3}}\\\midrule
     \multirow{3}{*}{KITTI} & No top-$k$ & \multicolumn{1}{r}{578.1}\\
      & 100 & \multicolumn{1}{r}{557.7}\\
      & 10 & \multicolumn{1}{r}{\textbf{482.9}}\\
     \bottomrule
     \label{tbl:analysis_k}
    \end{tabular}
  }
  \end{minipage}
  \hfill
  \begin{minipage}{.32\linewidth}
  \centering
  \subfloat[Effects of layers]
  {
    \begin{tabular}{ccr}
    \toprule
     Dataset & \multicolumn{1}{c}{Layers} & \multicolumn{1}{c}{FVD ($\downarrow$)} \\ 
     \midrule
     \multirow{2}{*}{BAIR} & 6 & \multicolumn{1}{r}{111.8} \\
      & 12 & \multicolumn{1}{r}{\textbf{99.3}} \\\midrule
     \multirow{2}{*}{KITTI} & 6 & \multicolumn{1}{r}{520.1}\\
      & 12 & \multicolumn{1}{r}{\textbf{482.9}}\\
     \bottomrule
     \label{tbl:analysis_layers}
    \end{tabular}
  }
  \end{minipage}
  \hfill
  \begin{minipage}{.32\linewidth}
  \centering
  \subfloat[Effects of $m$]
  {
    \begin{tabular}{ccc}
    \toprule
     Dataset & \multicolumn{1}{c}{$m$} & \multicolumn{1}{c}{FVD ($\downarrow$)} \\ 
     \midrule
     \multirow{4}{*}{KITTI} & 0 & \multicolumn{1}{r}{980.1}\\
      & 2 & \multicolumn{1}{r}{497.0}\\
      & 4 & \multicolumn{1}{r}{\textbf{482.9}}\\
      & 8 & \multicolumn{1}{r}{523.4}\\
     \bottomrule
     \label{tbl:analysis_m}
    \end{tabular}
  }
  \end{minipage}
  \hfill
  \label{tbl:analysis}
\end{table}

\paragraph{Results.}
Table~\ref{tbl:quantitative} shows the performances of our method and baselines on test sets of BAIR Robot Pushing and KITTI driving dataset.
We observe that our model achieves competitive or superior performance to state-of-the-art methods with large end-to-end networks, \textit{e.g.,} HARP outperforms FitVid with 302M parameters on KITTI driving dataset.
Our model successfully extrapolates to unseen number of future frames (\textit{i.e.,} 25) instead of 10 future frames used in training on KITTI dataset.
This implies that transformer-based video prediction models can also predict arbitrary number of frames at inference time. 
In the case of BAIR dataset, HARP achieves the similar performance of FitVid with 302M parameters, even though our method only requires 89M parameters.

\paragraph{Analysis.} 
We investigate how the top-$k$ sampling, number of layers, and magnitude $m$ of data augmentation affect the performance.
Table~\ref{tbl:analysis_k} shows that smaller $k$ leads to better performance, implying that the proposed top-$k$ sampling is effective for improving the performance by discarding rare discrete codes that might degrade the prediction quality at inference time.
As shown in Table~\ref{tbl:analysis_layers}, we observe that more layers leads to better performance on BAIR dataset, which implies our model can be further improved by scaling up the networks.
Finally, we find that (i) data augmentation on KITTI dataset is important for achieving strong performance, similar to the observation of \citet{babaeizadeh2021fitvid}, and (ii) too aggressive augmentation leads to worse performance.

\section{Discussion}
In this work, we present HARP that employs a high-fidelity image generator for predicting high-resolution future frames, and achieves competitive performance to state-of-the-art video prediction methods with large end-to-end networks.
We also demonstrate that HARP can leverage the image generator pre-trained on a wide range of natural images for video prediction, similar to the approach in the context of video synthesis \citep{tian2021good}.
We hope this work inspires more investigation into leveraging recently developed pre-trained image generators \citep{oord2017neural,chen2020generative,esser2021taming} for high-fidelity video prediction.

\begin{figure} [t] \centering
\subfloat[RoboNet]
{
\includegraphics[width=0.48\textwidth]{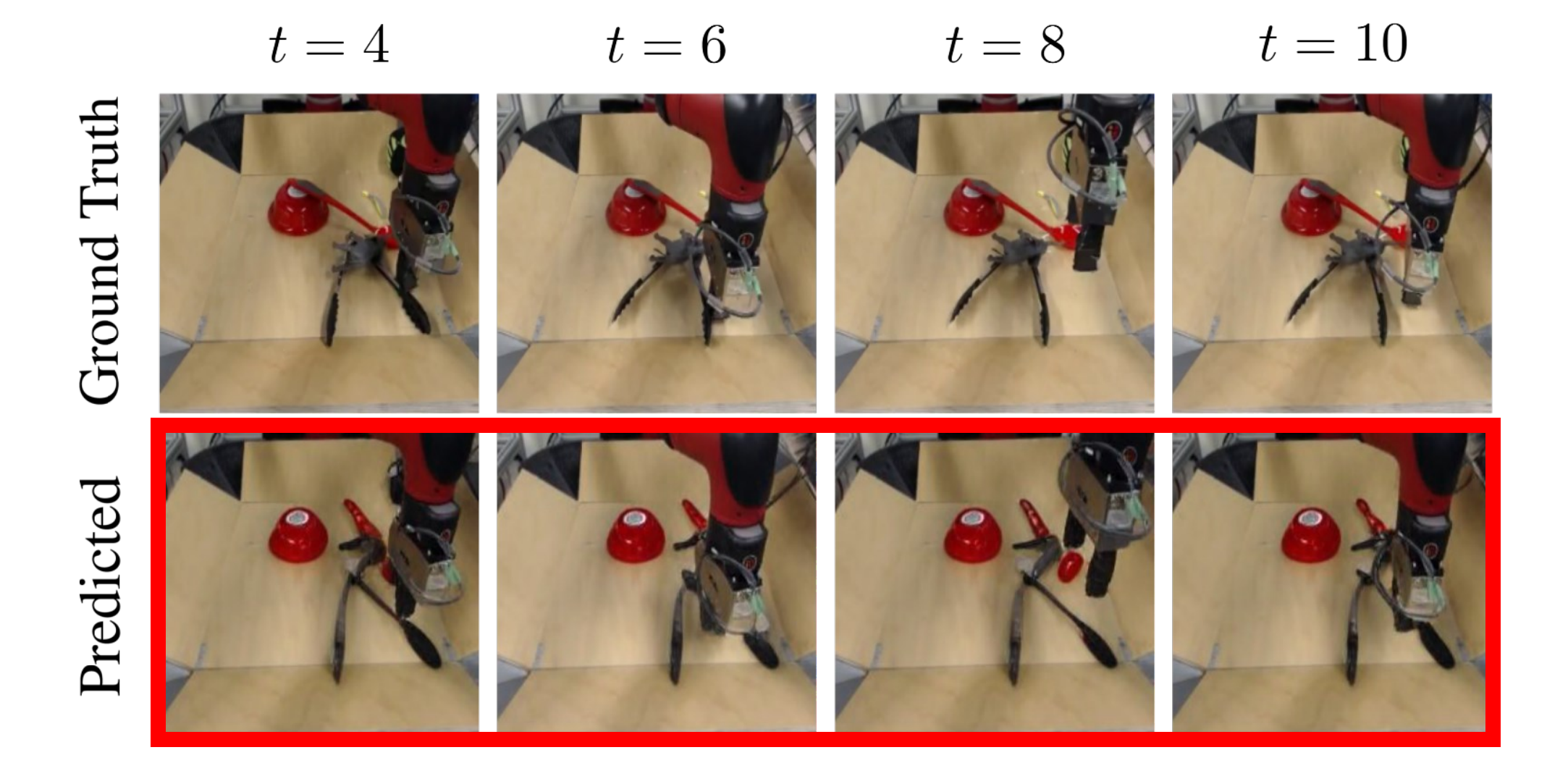}
\label{fig:failure_case_robonet}
} 
\subfloat[Kinetics-600]
{
\includegraphics[width=0.48\textwidth]{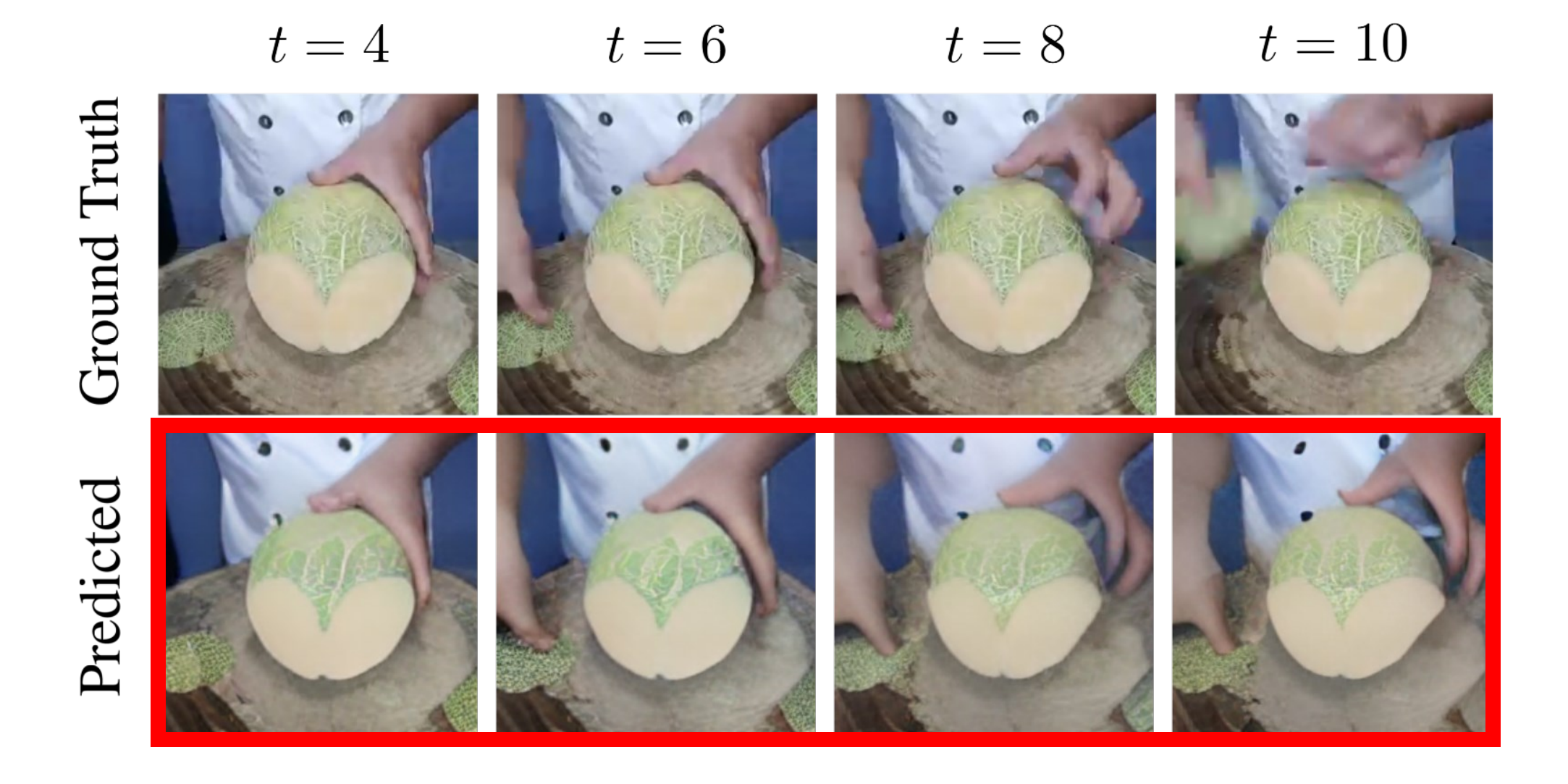}
\label{fig:failure_case_kinetics}
}
\caption{
Failure cases in our experiments. (a) Interaction with the objects is ignored. (b) The model repeats the first frame while a person is moving right in the ground-truth frames.
}
\label{fig:failure_case}
\end{figure}

Finally, we report the failure cases of video prediction with HARP and discuss the possible extensions to resolve the issue. A common failure case for video prediction on RoboNet dataset is ignoring the interaction between a robot arm and objects. 
For example, in Figure~\ref{fig:failure_case_robonet}, our model ignores the objects and only predicts the movement of a robot arm.
On the other hand, common failure case for Kinetics-600 is a degenerate video prediction, where a model just repeats the conditioning frame without predicting the future, as shown in Figure~\ref{fig:failure_case_kinetics}.
These failure cases might be resolved by training more larger networks similar to the observation in the field of natural language processing, \textit{e.g.,} GPT-3 \citep{brown2020language}, or might necessitate a new architecture for addressing the complexity of training autoregressive latent prediction models on video datasets. 

\section{Acknowledgements}
We would like to thank Jongjin Park, Wilson Yan, and Sihyun Yu for helpful discussions.
We also thank Cirrascale Cloud Services\footnote{\url{https://cirrascale.com}} for providing compute resources.
This work is supported by Institute of Information \& communications Technology Planning \& Evaluation (IITP) grant funded by the Korea government (MSIT) (No.2021-0-02068, Artificial Intelligence Innovation Hub; No.2019-0-00075, Artificial Intelligence Graduate School Program (KAIST)), Center for Human Compatible AI (CHAI), ONR N00014-21-1-2769, the Darpa RACER program, and the Hong Kong Centre for Logistics Robotics, BMW.

\bibliographystyle{arxiv}
\bibliography{arxiv}

\end{document}